# Entropy Computation of Document Images in Run-Length Compressed Domain

P. Nagabhushan[#2], Mohammed Javed[#1], B.B. Chaudhuri[*3]

[#]*Department of Studies in Computer Science*
*University of Mysore, Mysore-570006, India*
[1]javedsolutions@gmail.com
[2]pnagabhushan@hotmail.com

[*]*Computer Vision and Pattern Recognition Unit*
*Indian Statistical Institute, Kolkata-700108, India*
[3]bbc@isical.ac.in

*Abstract-* **Compression of documents, images, audios and videos have been traditionally practiced to increase the efficiency of data storage and transfer. However, in order to process or carry out any analytical computations, decompression has become an unavoidable pre-requisite. In this research work, we have attempted to compute the entropy, which is an important document analytic directly from the compressed documents. We use Conventional Entropy Quantifier (CEQ) and Spatial Entropy Quantifiers (SEQ) for entropy computations [1]. The entropies obtained are useful in applications like establishing equivalence, word spotting and document retrieval. Experiments have been performed with all the data sets of [1], at character, word and line levels taking compressed documents in run-length compressed domain. The algorithms developed are computational and space efficient, and results obtained match 100% with the results reported in [1].**

*Keywords-* Entropy, run-length compression, compressed documents, compressed domain processing

## I. INTRODUCTION

With the rapid growth of digital libraries and e-governance applications, the area of Document Image Analysis (DIA) has reached its zenith, which has led to drastic accretion in the volume of the data to be processed. Normally, any raw document image occupies large storage space and requires excessive bandwidth for communication. It is therefore useful if the documents are archived and transmitted in the compressed form. For further computer processing, necessarily the image is at first decompressed which means additional computational cost. An alternate and challenging proposal could be to perform operations directly on the compressed data, which may be called *compressed domain processing*.

PackBits, CCITT Group 3 (T.4)[2], CCITT Group 4 (T.6)[3] and JBIG are some of the popular algorithms supported by BMP, PDF, TIFF image formats to compress binary documents. Run Length Encoding (RLE) and its variant algorithms constitute backbone of these compression methods. Table-I, gives the description of horizontally compressed binary data using RLE technique. The compressed data consists of alternate columns of number of runs of 0 and 1 identified as odd columns (1, 3, 5, ...) and even columns (2, 4, 6, ..) respectively. Entropy computation which is an important document analytic used in establishing equivalence, retrieval, target spotting can be carried out easily in run-length compressed domain.

The entropy or energy content [4] of an image is a measure of the degree of randomness in the image, which is mathematically expressed as

$$H = \sum_{j=1}^{k} P(a_j) \log(P(a_j))$$

where, $a_1, a_2, a_3...a_k$ are a set of events associated with probabilities $p(a_1), p(a_2), p(a_3)...p(a_k)$. The concept of entropy [5] has many applications in the fields like image processing, pattern recognition, and machine learning.

TABLE I: Description of run-length compressed binary data

| Line | Binary image Data | 1 | 2 | 3 | 4 | 5 |
|---|---|---|---|---|---|---|
| 1: | 00000000000000 | 14 | 0 | 0 | 0 | 0 |
| 2: | 00110000111110 | 2 | 2 | 4 | 5 | 1 |
| 3: | 01111000111110 | 1 | 4 | 3 | 5 | 1 |
| 4: | 01111000111110 | 1 | 4 | 3 | 5 | 1 |
| 5: | 01111000111110 | 1 | 4 | 3 | 5 | 1 |
| 6: | 00110000000000 | 2 | 2 | 10 | 0 | 0 |
| 7: | 10000000000000 | 0 | 1 | 13 | 0 | 0 |
| 8: | 10000000000000 | 0 | 1 | 13 | 0 | 0 |
| 9: | 00100001111100 | 2 | 1 | 4 | 5 | 2 |
| 10: | 01110001111100 | 1 | 3 | 3 | 5 | 2 |
| 11: | 01111001111100 | 1 | 4 | 2 | 5 | 2 |
| 12: | 01111100000000 | 1 | 5 | 8 | 0 | 0 |
| 13: | 00000000000000 | 14 | 0 | 0 | 0 | 0 |

The authors in [1] use the entropy, which measures the energy contents distributed along the transitions in the document image, to analyze the contents of the text through the structure of the component. The entropy computations presented is the extension to the work demonstrated on uncompressed document images by [1]. In this paper, we report the results of implementing document analysis in run-length compressed domain.

The authors [1] compute entropy both in horizontal and vertical directions. But in case of horizontally compressed data, the vertical information becomes difficult to trace and hence is computationally expensive. Nevertheless, in this study we discuss entropy computations in both the directions, but in order to explore the computational efficiency of our proposed methods, we limit our experiment entropy calculations only to horizontal direction. With the experimental results obtained, it may be possible that many applications like equivalence detection, word-spotting and retrieval could be accomplished using horizontal entropy alone. Rest of the paper is organized as follows, section-2 describes the state-of-the-art techniques in the area of compressed domain, section-3 introduces the two entropy quantifiers, section-4 presents the experimental results and some discussions, section-5 highlights time complexity issues and finally section-6 concludes the paper with a brief summary.

## II. RELATED WORK

A run is a sequence of pixels having similar value and the number of such pixels is length of the run. The representation of the image pixels in the form of sequence of run values (*Pi*) and its lengths (*Li*) is run-length encoding. It can be mathematically represented as (*Pi, Li*). The earliest idea of operating directly on run-length encoded compressed documents can be related to the work of [6] and [7]. RLE, a simple compression method was first used for coding pictures [8] and television signals [9]. There are several efforts in the direction of directly operating on document images in compressed domain. Operations like image rotation [10], [11], connected component extraction [12], [13], skew detection [14], [15], [16], page layout analysis [16], [17], [18], [19], [20], [21], bar code detection [22] are reported in the literature related to run-length information processing. There are also some initiatives in finding document similarity [23], [24], equivalence [25] and retrieval [26].

One of the recent work using run-length information is to perform morphological related operations [27]. In most of the works, they use either run-length information from the uncompressed image or do some decoding to perform operations. But their computational cost, efficiency and scope has not been validated with large datasets. Nevertheless, we have initiated this research of which few of our works on feature extraction and segmentation using compressed data can be obtained from [28] and [29].

In the literature, entropy information extracted from uncompressed documents have been utilized for compression [30], [31], equivalence detection [1]. However, our research involves extraction of entropy from compressed data to analyze the contents directly from the compressed documents, which has the potential to facilitate many efficient applications in the area of image processing and pattern recognition.

## III. ENTROPY QUANTIFIERS

In this research work, we use the concept of entropy quantifiers proposed by [1] in the form of Conventional Entropy Quantifier (CEQ) and Sequential Entropy Quantifier (SEQ) for compressed documents. The details regarding the idea, motivation and formulation of CEQ and SEQ can be obtained in [1]. In general, CEQ measures the energy contribution of each row by considering the probable occurrence of +ve and -ve transitions among the total number of pixels in that row. However, in compressed data, this information is obtained by counting the number of even and odd columns except the first column, for $0-1$ and $1-0$ transitions respectively as shown in Table-I. Such an entropy is mathematically expressed as

$$\text{For CEQ, } E(t) = p * \log(\tfrac{1}{p}) + (1-p) * \log(\tfrac{1}{(1-p)})$$

where $t$ is the transition from $0-1$ and $1-0$, $E(t)$ is the entropy. For the compressed data, $p$ is calculated based on the number of even or odd non-zero columns in each row divided by the sum of all the runs of that particular row. Thus if $p$ is the probable occurrence of transition in each row, then $1-p$ is the probable non-occurrence of transition.

Entropy of even $(0-1)$ and odd $(1-0)$ columns are computed independently and termed as $E^+(t)$ and $E^-(t)$ respectively, and total entropy of each row is the summation of $E^+(t)$ and $E^-(t)$ as shown below,

$$E(t) = E^+(t) + E^-(t)$$

Similar entropy computations for all the rows results in total horizontal entropy $Eh(t)$. CEQ measures the absolute energy contributed by the structure of the component irrespective of the position of occurrence of transitions and entropy remains unchanged in case of characters like *u* and *v*, words like *intended* and *indented* as shown by [1]. In order to overcome this drawback, the authors propose a new entropy measure based on the position of occurrence of energy termed as SEQ, that is briefly discussed below.

Spatial Entropy Quantifier analyzes the component by measuring the entropy at the position of occurrence of the transitions. The probable occurrence $p$ in CEQ is replaced by the position parameter *pos*, which indicates the position of transition point in horizontal direction. For compressed data, the transition position at each column is obtained by the addition of run values of all the previous columns and followed by one increment.

If the transition occurs between two columns $C_{\beta 1}$ and $C_{\beta 2}$ in row $r_\alpha$ then corresponding row entropy is formulated as:

$$E(\beta) = \frac{r_\alpha}{m}\left(\frac{pos}{n} * \log\frac{n}{pos} + \left(m - \frac{pos}{n}\right) * \log\frac{m}{m*n - pos}\right)$$

where $\beta = 1, ..., m$. Total entropy for each row is the summation of entropy at even and odd columns represented as $E^+(\beta)$ and $E^-(\beta)$ respectively given as

$$E(\beta) = E^+(\beta) + E^-(\beta)$$

The total horizontal entropy $E_h(\beta)$ is the summation of $E(\beta)$ of all the rows of compressed data.

Entropy computations in vertical direction is not straight forward. This is because the vertical information is not directly available in the compressed representation. So, we obtain the transition points by pulling-out the run values from all the rows simultaneously using first two columns from compressed data shown in Table-I. In presence of zero run values in both the columns, the runs on the right are shifted to two positions leftwards. Thus for every pop operation, the run value is decremented by 1 and if the popped-out element is from first column then the transition value is 0 otherwise 1. This process is repeated for all the rows generating a sequence of column transitions from the compressed file which may be called *virtual decompression*. Table-II shows virtual decompression for first 10 passes using first four lines of compressed data of Table-I. Thus, collecting all the popped-out column transitions, CEQ and SEQ entropies are computed using the formulas proposed by [1] respectively as $E_v(t)$ and $E_v(\beta)$. The time complexity issues and other analysis can be obtained from [28].

TABLE II: Virtual decompression of run-length compressed data

| Pass | Line | Popped Transitions | 1 | 2 | 3 | 4 | 5 | Status |
|---|---|---|---|---|---|---|---|---|
| Start | 1: |  | 14 | 0 | 0 | 0 | 0 |  |
|  | 2: |  | 2 | 2 | 4 | 5 | 1 |  |
|  | 3: |  | 1 | 4 | 3 | 5 | 1 |  |
|  | 4: |  | 1 | 4 | 3 | 5 | 1 |  |
| 1 | 1: | 0 | 13 | 0 | 0 | 0 | 0 | pop |
|  | 2: | 0 | 1 | 2 | 4 | 5 | 1 | pop |
|  | 3: | 0 | 0 | 4 | 3 | 5 | 1 | pop |
|  | 4: | 0 | 0 | 4 | 3 | 5 | 1 | pop |
| 2 | 1: | 0 | 12 | 0 | 0 | 0 | 0 | pop |
|  | 2: | 0 | 0 | 2 | 4 | 5 | 1 | pop |
|  | 3: | 1 | 0 | 3 | 3 | 5 | 1 | pop |
|  | 4: | 1 | 0 | 3 | 3 | 5 | 1 | pop |
| 3 | 1: | 0 | 11 | 0 | 0 | 0 | 0 | pop |
|  | 2: | 1 | 0 | 1 | 4 | 5 | 1 | pop |
|  | 3: | 1 | 0 | 2 | 3 | 5 | 1 | pop |
|  | 4: | 1 | 0 | 2 | 3 | 5 | 1 | pop |
| 4 | 1: | 0 | 10 | 0 | 0 | 0 | 0 | pop |
|  | 2: | 1 | 0 | 0 | 4 | 5 | 1 | pop |
|  | 3: | 1 | 0 | 1 | 3 | 5 | 1 | pop |
|  | 4: | 1 | 0 | 1 | 3 | 5 | 1 | pop |
| 5 | 1: | 0 | 9 | 0 | 0 | 0 | 0 | pop |
|  | 2: | 0 | 3 | 5 | 1 |  |  | shift-pop |
|  | 3: | 1 | 0 | 0 | 3 | 5 | 1 | pop |
|  | 4: | 1 | 0 | 0 | 3 | 5 | 1 | pop |
| 6 | 1: | 0 | 8 | 0 | 0 | 0 | 0 | pop |
|  | 2: | 0 | 2 | 5 | 1 |  |  | pop |
|  | 3: | 0 | 2 | 5 | 1 |  |  | shift-pop |
|  | 4: | 0 | 2 | 5 | 1 |  |  | shift-pop |
| 7 | 1: | 0 | 7 | 0 | 0 | 0 | 0 | pop |
|  | 2: | 0 | 1 | 5 | 1 |  |  | pop |
|  | 3: | 0 | 1 | 5 | 1 |  |  | pop |
|  | 4: | 0 | 1 | 5 | 1 |  |  | pop |
| 8 | 1: | 0 | 6 | 0 | 0 | 0 | 0 | pop |
|  | 2: | 0 | 0 | 5 | 1 |  |  | pop |
|  | 3: | 0 | 0 | 5 | 1 |  |  | pop |
|  | 4: | 0 | 0 | 5 | 1 |  |  | pop |
| 9 | 1: | 0 | 5 | 0 | 0 | 0 | 0 | pop |
|  | 2: | 1 | 0 | 4 | 1 |  |  | pop |
|  | 3: | 1 | 0 | 4 | 1 |  |  | pop |
|  | 4: | 1 | 0 | 4 | 1 |  |  | pop |
| 10 | 1: | 0 | 4 | 0 | 0 | 0 | 0 | pop |
|  | 2: | 1 | 0 | 3 | 1 |  |  | pop |
|  | 3: | 1 | 0 | 3 | 1 |  |  | pop |
|  | 4: | 1 | 0 | 3 | 1 |  |  | pop |

## IV. EXPERIMENTAL DETAILS

In this research work, experiments are conducted on various components at character, word and line level similar to the work demonstrated in [1], where it is on uncompressed documents. They have considered entropy values in both horizontal and vertical directions. Since the computation of vertical entropy is expensive in case of compressed data as shown in Table-III for sample document in Fig-1. However, in this study we limit our experiments to horizontal entropy. This is because, we intend to explore horizontal entropy for applications of document equivalence, word-spotting, etc as an extension to this research work.

TABLE III: Vertical entropy computation time in *seconds* for CEQ

| CEQ | Compressed Data | Decompressed Data |
|---|---|---|
| Sample Document | 0.5745 | 0.0513 |

The energy features defined in the above section are retermed as $F1$, $F2$ and $F3$ where, $F1$ represents $E^+(t)$ or $E^+(\beta)$, $F2$ represents $E^-(t)$ or $E^-(\beta)$ respectively in case of CEQ or SEQ, and $F3$ is the summation of $F1$ and $F2$ giving the total entropy. Also feature $F3$ is used for distance calculations.

CHARACTER LEVEL: Even though every character has unique structure, they may be similar in resemblance like *O* and 0, 1 and *l*, and may go unnoticed and get miss recognized as said in [1]. On the other hand, we have done experimentations taking the same character with different font styles such as Times New Roman (TNR) and Arial. The features *F*1, *F*2, and *F*3 have been extracted using both CEQ and SEQ entropy models as shown below by Table-IV and Table-V.

TABLE IV: Feature extracted through CEQ

| Font Style | F1 | F2 | F3 |
|---|---|---|---|
| TNR C | 1.91 | 1.91 | 3.82 |
| Arial C | 1.86 | 1.86 | 3.71 |
| TNR S | 2.27 | 2.27 | 4.54 |
| Arial S | 1.98 | 1.98 | 3.96 |
| TNR D | 2.02 | 2.02 | 4.01 |
| Arial D | 2.23 | 2.23 | 4.48 |

TABLE V: Feature extracted through SEQ

| Font Style | F1 | F2 | F3 |
|---|---|---|---|
| TNR C | 72.24 | 70.78 | 143.02 |
| Arial C | 87.65 | 86.10 | 173.76 |
| TNR S | 80.44 | 78.43 | 158.87 |
| Arial S | 86.35 | 84.43 | 170.78 |
| TNR D | 90.67 | 88.85 | 179.52 |
| Arial D | 115.97 | 113.82 | 229.79 |

The distance matrix obtained taking entropy features extracted using SEQ model shows a clear discrimination, whereas CEQ fails to preserve. Distances computed with total entropy alone *F*3 using both CEQ and SEQ are shown below in Table-VI and Table-VII, respectively. The distances between the components with SEQ show higher discrimination values compared to CEQ.

TABLE VI: Distance matrix using CEQ

| Sample | TNR C | Arial C | TNR S | Arial S | TNR D | Arial D |
|---|---|---|---|---|---|---|
| TNR C | 0 | 0.10 | 0.73 | 0.14 | 0.23 | 0.66 |
| Arial C | 0.10 | 0 | 0.83 | 0.25 | 0.33 | 0.76 |
| TNR S | 0.73 | 0.83 | 0 | 0.58 | 0.50 | 0.07 |
| Arial S | 0.14 | 0.25 | 0.58 | 0 | 0.08 | 0.52 |
| TNR D | 0.23 | 0.33 | 0.50 | 0.08 | 0 | 0.43 |
| Arial D | 0.66 | 0.76 | 0.07 | 0.52 | 0.43 | 0 |

WORD LEVEL: Word is a sequence of characters. There are many words made up of same character components but placed at different positions such as in words like *top* and *pot*, *was* and *saw*. The distance computations using CEQ for such words fails to make distinction because the number of transitions in a row remain same for such combinations, whereas SEQ which works based on position shows clear discrimination. Table-VIII and Table-IX show the results considering some combinations of English characters.

Experiments have been conducted by considering a single word like *tub* with all combinations of *t*, *u* and *b* as *but*, *utb*, *tbu*, *tub*, *ubt*, and *btu*. These words do not convey any meaning but for the purpose of testing the strength of the algorithm we have constructed these words and the distance between every pair of components using feature *F*3 have been tabulated below. CEQ fails completely in case like *tbu* and *ubt*, and shows almost zero discrimination for some cases like *utb* and *tub*, *tbu* and *tub*, where as SEQ shows clear discrimination among them as shown in Table-X and Table-XI respectively.

TABLE VII: Distance matrix using SEQ

| Sample | TNR C | Arial C | TNR S | Arial S | TNR D | Arial D |
|---|---|---|---|---|---|---|
| TNR C | 0 | 30.74 | 15.86 | 27.77 | 36.51 | 86.78 |
| Arial C | 30.74 | 0 | 14.89 | 2.97 | 5.76 | 56.03 |
| TNR S | 15.86 | 14.89 | 0 | 11.91 | 20.65 | 70.92 |
| Arial S | 16.56 | 2.97 | 11.91 | 0 | 8.736 | 59.006 |
| TNR D | 36.51 | 5.76 | 20.65 | 8.74 | 0 | 50.27 |
| Arial D | 86.78 | 56.03 | 70.92 | 59.00 | 50.27 | 0 |

TABLE VIII: Discrimination matrix using CEQ

| CEQ | F1 | F2 | F3 | Distance |
|---|---|---|---|---|
| impunity | 2.069257788 | 2.069257788 | 4.138515575 | 0.000248 |
| impurity | 2.069382067 | 2.069382067 | 4.138764134 | |
| intended | 2.077617901 | 2.077617901 | 4.155235803 | 0 |
| indented | 2.077617901 | 2.077617901 | 4.155235803 | |
| message | 1.753523895 | 1.753523895 | 3.507047789 | 0.022997 |
| massage | 1.765022305 | 1.765022305 | 3.53004461 | |

TABLE IX: Discrimination matrix using SEQ

| CEQ | F1 | F2 | F3 | Distance |
|---|---|---|---|---|
| impunity | 630.8360158 | 628.6150706 | 1259.451086 | 60.23 |
| impurity | 600.6986346 | 598.5217363 | 1199.220371 | |
| intended | 705.1652411 | 702.3179657 | 1407.483207 | 0.49 |
| indented | 705.4075819 | 702.5628731 | 1407.970455 | |
| message | 320.4200367 | 318.8765993 | 639.296636 | 9.202 |
| massage | 325.0469003 | 323.4525886 | 648.4994889 | |

TABLE X: Distance matrix using CEQ

| Sample | but | utb | tbu | tub | ubt | btu |
|---|---|---|---|---|---|---|
| but | 0 | 0.35 | 0.22 | 0.29 | 0.22 | 0.17 |
| utb | 0.35 | 0 | 0.12 | 0.05 | 0.12 | 0.18 |
| tbu | 0.22 | 0.12 | 0 | 0.07 | 0 | 0.05 |
| tub | 0.29 | 0.05 | 0.07 | 0 | 0.07 | 0.12 |
| ubt | 0.22 | 0.12 | 0 | 0.07 | 0 | 0.05 |
| btu | 0.17 | 0.18 | 0.05 | 0.12 | 0.05 | 0 |

TABLE XI: Distance matrix using SEQ

| Sample | but | utb | tbu | tub | ubt | btu |
|---|---|---|---|---|---|---|
| but | 0 | 43.35 | 46.14 | 32.57 | 34.43 | 35.82 |
| utb | 43.35 | 0 | 2.787 | 8.92 | 1.19 | 7.53 |
| tbu | 46.14 | 2.79 | 0 | 11.70 | 1.60 | 10.31 |
| tub | 32.57 | 8.92 | 11.70 | 0 | 10.10 | 1.39 |
| ubt | 34.54 | 1.19 | 1.60 | 10.10 | 0 | 8.71 |
| btu | 35.82 | 7.53 | 10.31 | 1.39 | 8.71 | 0 |

LINE LEVEL: Line is a set of words. Experiments have been conducted to distinguish a line with different word combinations as shown below. Consider the lines shown below,

Line-1: Radha went to school in the afternoon.
Line-2: in the afternoon Radha went to school.
Line-3: went to school Radha in the afternoon.
Line-4: went to school in the afternoon Radha.

The distance matrix using CEQ and SEQ are tabulated in Table-XII and Table-XIII respectively.

TABLE XII: Distance matrix for line using CEQ

| Sample | Line-1 | Line-2 | Line-3 | Line-4 |
|---|---|---|---|---|
| Line-1 | 0 | 0.02 | 0 | 0 |
| Line-2 | 0.02 | 0 | 0.02 | 0.02 |
| Line-3 | 0 | 0.02 | 0 | 0 |
| Line-4 | 0 | 0.02 | 0 | 0 |

TABLE XIII: Distance matrix for line using SEQ

| Sample | Line-1 | Line-2 | Line-3 | Line-4 |
|---|---|---|---|---|
| Line-1 | 0 | 43.61 | 6.49 | 11.32 |
| Line-2 | 43.61 | 0 | 37.12 | 32.29 |
| Line-3 | 6.49 | 37.12 | 0 | 4.83 |
| Line-4 | 11.32 | 32.29 | 4.83 | 0 |

The distance matrix using CEQ shows almost no discrimination and it fails to recognize the words being shuffled in the same sentence. But SEQ shows clear discrimination between the samples considered.

V. COMPLEXITY ISSUES AND DISCUSSIONS

The proposed algorithms in run-length compressed domain can be theoretically compared with that of conventional method. The algorithm works in two stages,
*Stage* 1: Search for transition points
*Stage* 2: Entropy Computations

In order to analyze the efficiency and time complexity of working with compressed and decompressed documents, consider a document of size $m \times n$ and its compressed version of size $m \times k$ where $k<n$. The best case to the algorithm is blank image or image made up of only zero value pixels. In the worst case, the image will have maximum number of transitions ($n-1$ in each row). CEQ has the best case of $m \times n$ search and zero entropy computation for decompressed and $m$ search and zero entropy computation for compressed image. On the other hand, for the worst case, the decompressed image has $(m \times n)$ search and $m$ entropy computations and for a compressed image $m \times (k-1)$ search and $m$ entropy computations. In SEQ, the best case for an decompressed image is $(m \times n)$ search and zero entropy computation and for a compressed image $(m \times 1)$ search and zero entropy computation. The worst case for decompressed image is $(m \times n)$ search $m \times (n-1)$ entropy computations, whereas for compressed image $m \times (k-1)$ search $m \times (k-1)$ entropy computations. Therefore the time complexity of entropy computations for run-length compressed data is always less than that of decompressed data.

Overall, in this research work, the entire data sets of [1] have been experimented with the proposed algorithms on compressed documents working in run-length compressed domain. The entropy calculations obtained from our experiments match 100% to that of uncompressed documents as shown experimentally in [28]. This is because the horizontal runlength compression is lossless and also preserves all the row transition points. The proposed algorithms for entropy computation do not require decompression and in fact works computationally faster in compressed domain. Entropy computation is done at character, word, line and document level using both compressed (comp) and uncompressed (uncomp) documents and execution time is noted as shown in Table-XIV.

TABLE XIV: Horizontal entropy time complexity analysis (in *seconds*)

| Sample | CEQ | | | SEQ | | |
|---|---|---|---|---|---|---|
| | Comp | uncomp | Ratio | Comp | uncomp | Ratio |
| Character | 0.0041 | 0.0065 | 1.59 | 0.00007 | 0.0066 | 94.28 |
| Word | 0.0046 | 0.0062 | 1.35 | 0.00016 | 0.0065 | 40.62 |
| Line | 0.0042 | 0.0068 | 1.62 | 0.00057 | 0.0067 | 11.75 |
| Document | 0.0213 | 0.16 | 7.51 | 0.0479 | 0.1741 | 3.63 |

The ratio of execution time for uncompressed and compressed document gives the increase in execution speed. As expected we see working in compressed domain is computationally very efficient. This is due to the fact that runlength compressed data is very compact which needs fewer computations as shown in Table-I. Also, the buffer space occupied by the compressed data is comparatively less than that of decompressed data. In order to illustrate this fact, consider the scanned document in Fig-1, the CEQ and SEQ computations in compressed domain are respectively 7.51 and 3.63 times faster than conventional method as shown in Table-XIV. Therefore, all these facts establish the superiority of our proposed model.

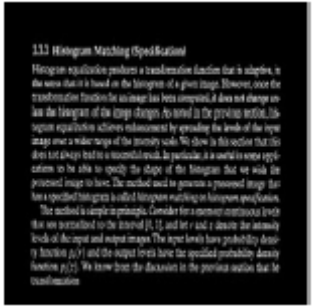

Fig. 1: Scanned sample document

## VI. CONCLUSION

A method to compute the entropy directly and efficiently from the run-length compressed documents has been demonstrated with the help of two entropy quantifiers namely Conventional Entropy Quantifier (CEQ) and Spatial Entropy Quantifier (SEQ). The entropy values obtained are similar to that of decompressed document. The proposed model is superior in time and space to that of conventional methods. Therefore, the entropy calculations demonstrated in this research work can be used for document understanding in run-length compressed domain, which can facilitates larger efficient applications like document equivalence and retrieval.